*Article*

# Neural Modulation Alteration to Positive and Negative Emotions in Depressed Patients: Insights from fMRI Using Positive/Negative Emotion Atlas

**Yu Feng [1], Weiming Zeng [1,*], Yifan Xie [1], Hongyu Chen [1], Lei Wang [1], Yingying Wang [1], Hongjie Yan [2], Kaile Zhang [3], Ran Tao [3], Wai Ting Siok [3] and Nizhuan Wang [3,*]**

[1] Lab of Digital Image and Intelligent Computation, College of Information Engineering, Shanghai Maritime University, Shanghai 201306, China; fyuchn@163.com (Y.F.); x1_shiny@163.com (Y.X.); hongychen676@gmail.com (H.C.); sayhiwl@163.com (L.W.); 202230310057@stu.shmtu.edu.cn (Y.W.)
[2] Department of Neurology, Affiliated Lianyungang Hospital of Xuzhou Medical University, Lianyungang 222002, China; yanhjns@gmail.com (H.Y.)
[3] Department of Chinese and Bilingual Studies, The Hong Kong Polytechnic University, Hong Kong, China; kaile-keller.zhang@polyu.edu.hk (K.Z.); ran.tao@polyu.edu.hk (R.T.); wai-ting.siok@polyu.edu.hk (W.T.S.)
* Correspondence: zengwm86@163.com (W.Z.); wangnizhuan1120@gmail.com (N.W.)

**Abstract:** Background: Although it has been noticed that depressed patients show differences in processing emotions, the precise neural modulation mechanisms of positive and negative emotions remain elusive. FMRI is a cutting-edge medical imaging technology renowned for its high spatial resolution and dynamic temporal information, making it particularly suitable for the neural dynamics of depression research. Methods: To address this gap, our study firstly leveraged fMRI to delineate activated regions associated with positive and negative emotions in healthy individuals, resulting in the creation of the positive emotion atlas (PEA) and the negative emotion atlas (NEA). Subsequently, we examined neuroimaging changes in depression patients using these atlases and evaluated their diagnostic performance based on machine learning. Results: Our findings demonstrate that the classification accuracy of depressed patients based on PEA and NEA exceeded 0.70, a notable improvement compared to the whole-brain atlases. Furthermore, ALFF analysis unveiled significant differences between depressed patients and healthy controls in eight functional clusters during the NEA, focusing on the left cuneus, cingulate gyrus, and superior parietal lobule. In contrast, the PEA revealed more pronounced differences across fifteen clusters, involving the right fusiform gyrus, parahippocampal gyrus, and inferior parietal lobule. Conclusions: These findings emphasize the complex interplay between emotion modulation and depression, showcasing significant alterations in both PEA and NEA among depression patients. This research enhances our understanding of emotion modulation in depression, with implications for diagnosis and treatment evaluation.

**Keywords:** fMRI; positive emotion; negative emotion; depression; SVM; ALFF

## 1. Introduction

Depression, also known as depressive disorder, is a serious mental illness characterized by elevated prevalence, frequent recurrence, significant suicide-related mortality, and a substantial disease burden [1]. The fundamental symptoms of depression encompass heightened negative emotions and an absence of positive affect [2]. Patients with depression often exhibit pronounced negative emotions such as sadness, anxiety, irritability, and self-blame. Positive emotions typically offer benefits, distinct from negative emotions. Nevertheless, in patients with depression, positive emotions may exhibit complex features. Some patients with depression may have a diminished response to positive emotions, posing challenges for them to embrace sensations like happiness and satisfaction [3]. Conversely, under specific circumstances, patients with depression may exhibit intense positive emotional reactions. However, these moments of positivity are transient, swiftly overshadowed by a resurgence of depression [4]. Therefore, investigating the etiology and therapeutic mechanisms of depression remains a central research objective. Clinical practice primarily relies on drug treatment for depression, but approximately 30% of depressed patients exhibit poor responsiveness to medication [5]. Moreover, the variations in brain regions and the diverse treatment requisites of patients with different subtypes, severities, and accompanying symptoms are highly heterogeneous. Conventional diagnostic and therapeutic modalities encounter challenges in accurately evaluating the condition of these patients.

Recently, a plethora of neuroimaging modalities has emerged to investigate the structure and function of the human brain, with fMRI prominently positioned as a pivotal methodology. This technique efficiently captures fluctuations in blood oxygen level-dependent signals within the brain, providing valuable perspectives into the activity states of various brain regions and their interrelationships [6]. Given the temporal and spatial resolution of fMRI imaging, along with the recognition that abnormal manifestations of depression primarily arise from atypical activity and interactions within brain regions [7], an increasing number of researchers are turning to fMRI for the study of depression-related phenomena. For instance, Sheline et al. [8] employed fMRI to evaluate resting-state functional connectivity (RSFC) within cognitive control, default mode, and affective networks in individuals with depression. They found increased connectivity in all three networks, particularly with ipsilateral dorsomedial prefrontal cortex regions, compared to healthy controls. Seema and Shankapal [9] examined differential brain-activation patterns between depressed patients and healthy individuals during various music stimulation tasks using fMRI. Their results underscored significant activation within the anterior cingulate cortex, dorsolateral prefrontal cortex, and striatum in individuals with depression. Rubin-Falcone et al. [10] employed fMRI to study the neural correlates of emotional reactivity and emotion regulation during the viewing of emotionally salient images as predictors of treatment outcomes with Cognitive Behavioral Therapy (CBT) for major depressive disorder (MDD). Their results indicated that the neural correlates of emotional reactivity might possess stronger predictive power for CBT outcomes. Phan et al.'s [11] meta-analysis elucidates the differential effects of various emotions on brain-activation areas. The medial prefrontal cortex universally mediates emotion processing, while specific emotions engage distinct regions: fear primarily activates the amygdala, sadness correlates with the subgenual cingulate, visual stimuli stimulate the occipital cortex and amygdala, emotional recall and imagination involve the anterior cingulate and insula, and cognitively demanding emotional tasks primarily engage the anterior cingulate and insula. Murphy et al.'s [12] meta-analysis investigates the impact of different emotions on brain activity through the lens of left–right brain symmetry and asymmetry. Their findings indicate that approach emotions are associated with greater left-brain activity, particularly exhibiting significant asymmetry in anterior regions. Conversely, negative or withdrawal emotions demonstrate symmetry in left–right brain activity. To date, researchers have predominantly focused on aberrant alterations in brain structure, functional connectivity (FC), and neural activity in individuals with depression [13–15]. However, few studies have delved into the distinctions of relevant brain regions in depressed patients, particularly concerning alterations in regions associated with positive and negative emotions as observed in neurotypical individuals. Moreover, a dearth of comprehensive research exists in elucidating the changes and underlying mechanisms in specific brain regions closely linked to positive and negative emotions in depressed patients.

In current studies, the analysis and diagnosis of brain disorders generally rely on the universal template, which provides a basic framework for disease research but lack specificity when addressing specific diseases like depression [16]. Due to the intricate emotional fluctuations and individual diversities associated with depression, generic templates often fail to capture the unique emotional changes of patients, resulting in unsatisfactory diagnostic efficacy. Therefore, a depression-specific emotional template is essential to precisely capture the subtle differences in patients' emotional changes [17]. Additionally, the accuracy of traditional Region of Interest (ROI) level templates in depression diagnosis has been notably limited. In fMRI imaging, a voxel serves as the smallest unit of analysis, similar to a pixel in a two-dimensional image [18]. The fMRI-derived images are partitioned into abstract 3D grids, with each unit termed a voxel [19]. Given the registration process conducted according to distinct brain region templates, an indefinite number of voxels populate a brain region, allowing voxel-level investigations to offer a more nuanced analysis and potentially deeper insights into depression [20]. Despite their advantages, voxel-based fMRI analysis methods are susceptible to technical limitations like image registration and spatial transformation, which may affect the accuracy and reliability of results [21]. According to the Two-Factor Model of Affect [22], positive affect and negative affect constitute two primary and relatively independent dimensions of the affective structure, exhibiting significant differences in emotional regulation and neural mechanisms. These differences are not only reflected in individuals' emotional experiences but also impact brain activity. Therefore, this paper introduces the Positive Emotion Atlas (PEA) and Negative Emotion Atlas (NEA) as solutions to overcome these challenges. We anticipate that the PEA and NEA can play a crucial role in depression research and treatment, serving as supplementary diagnostic tools to aid physicians in the initial identification of individuals with depression. Through continuously monitoring the changes in patients' PEA and NEA throughout the treatment process, the effectiveness of the therapy can be assessed and the treatment plan can be adjusted accordingly. Furthermore, the establishment of PEA and NEA also aids in refining brain imaging data analysis methods. Integration with other brain imaging technologies like EEG allows for the fusion of multimodal data analysis, facilitating the construction of a more comprehensive and sophisticated analytical template.

In summary, two key questions regarding the relationship between negative/positive emotion modulation and depression remain unresolved. Firstly, the precise delineation of PEA and NEA is deficient, despite their significant activation in normal individuals in response to positive and negative emotional picture stimulation, respectively. Secondly, it is imperative to investigate the changes that occur in depressed patients under the PEA and NEA and determine whether they differ from those in healthy controls. To bridge these gaps, this study utilized fMRI technology to elucidate the activation regions associated with positive and negative emotions in healthy controls at both brain region and voxel levels, thereby constructing the PEA and NEA. Subsequently, a Support Vector Machine (SVM) classifier [23] was trained to distinguish between depression patients and healthy controls based on the entire brain and the constructed PEA and NEA. To enhance classification performance, the Cost-Sensitive Learning (CSL) [24] strategy was integrated into the SVM classifier. Further analysis utilizing Amplitude of Low-Frequency Fluctuation (ALFF) [25] was designed to reveal significant differences between depressed patients and healthy controls.

## 2. Materials and Methods

### *2.1. Data Acquisition*

#### 2.1.1. Positive/Negative Stimulus Task FMRI Dataset

When selecting participants for the emotional stimulus dataset, we considered the following factors: First, ensuring that the participants are in a normal mental state. Secondly, choosing participants of similar ages to minimize the impact of differences in social experience, psychological resilience, and perspective on the experimental results. Thirdly, considering the educational background of the participants, as different academic disciplines may influence the understanding of images. Finally, striving to maintain a balanced gender ratio to reduce the impact of gender on emotional responses. Taking all these factors into account, twenty-one participants (12 males, 9 females) with an average age of 23.65 ± 1.5 years were enrolled in the study, where part of participants was used in our previous study [2]. All participants possessed normal or corrected-to-normal visual acuity, maintained good health with no history of mental or serious physical illness, and were right-handed. Prior to the experiment, participants were fully informed of the study procedures and provided written informed consent in accordance with the guidelines approved by the Institutional Review Board (IRB) of East China Normal University (ECNU). This study adhered to ethical principles and guidelines.

The fMRI images were obtained during exposure to positive and negative emotional stimuli, followed by resting-state scans. Stimulus images were sourced from the International Affective Picture System (IAPS) [26], to ensure consistency and reliability. The experimental paradigm, as depicted in Figure 1, involved alternating periods of rest and emotional stimulation: rest–positive–rest–negative–rest–positive–rest–negative–rest–positive–rest–negative–rest. A 20 s resting period preceded the formal experiment, allowing subjects to stabilize their emotions. Each formal experiment lasted 240 s and comprised six blocks of 40 s each, alternating between task states (positive/negative emotional picture stimuli) and rest periods. Subjects viewed positive or negative emotional stimuli for 20 s during task blocks, randomly selected from the picture library. During rest periods, subjects lay flat with their heads fixed and were instructed to refrain from active thought while fixating on a white cross against a black background.

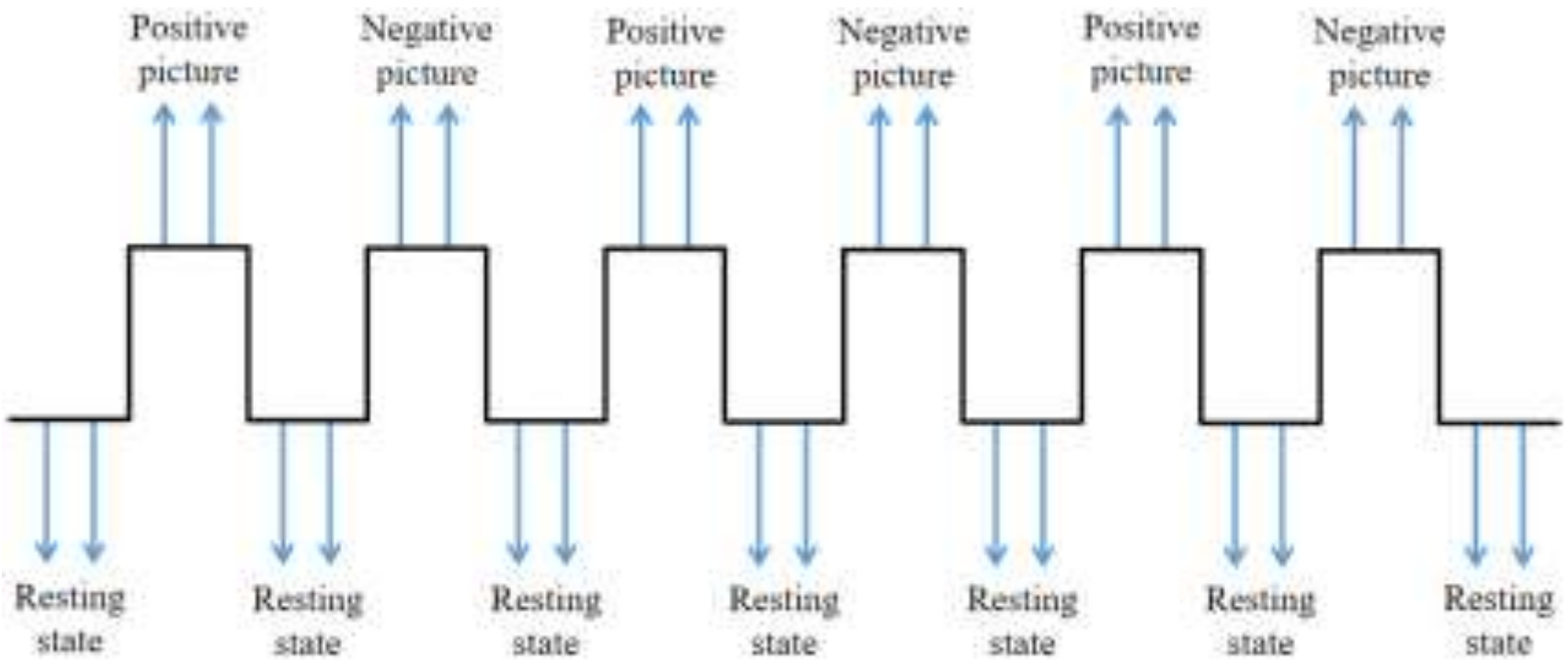

**Figure 1.** Schematic illustration of fMRI data-acquisition paradigm [27].

The fMRI data of 21 subjects were acquired at the Shanghai Key Laboratory of Magnetic Resonance at ECNU using a GE 3.0 Tesla MRI scanner. The imaging protocol employed a single-shot gradient echo planar imaging sequence comprising 33 slices, with a sensitivity acceleration factor of 2.0. Parameters included a repetition time (TR) of 2.0 s, a scan resolution of 64 × 64, an in-slice resolution of 3 mm × 3 mm, a slice thickness of 4 mm, and a slice interval of 1 mm.

### 2.1.2. Resting-State Depression FMRI Dataset

The depression dataset used in this study was obtained from the public dataset OpenNeuro "https://openneuro.org/ (accessed on 06 July 2023)", with Accession Number DS002748 [28]. This dataset complied with the Helsinki Declaration and the ethics board of the Research Institute of Molecular Biology and Biophysics in Novosibirsk. The original dataset comprised 51 depression patients and 21 healthy controls. The control group consisted of individuals who were deemed healthy, lacking psychotic disorders, severe neurological and somatic disorders (as confirmed by a neurologist), and having no contraindications for MRI. However, given the limited number of subtypes and the study was restricted to the category of depression, we excluded one patient with dysthymia (sub-10), one with persistent mood disorder (sub-30), and one unannotated case (sub-34). Additionally, during the data preprocessing phase, we found that the data signal quality of sub-06, sub-15, and sub-60 was poor, characterized by significant image noise.

Consequently, the data from these three subjects were also excluded. Ultimately, we selected resting-state fMRI data from 46 depression patients and 20 healthy controls from the original dataset, the details are shown in Table 1.

**Table 1.** Demographic and clinical characteristics of the groups involved in the study; M: males; F: females; SD: standard deviation; MADRS: Montgomery–Asberg depression rating scale; BDI: Beck depression inventory; ZSRDS: Zung self-rating depression scale.

| Group | Sex | Age, Mean ± SD | IQ, Mean ± SD | MADRS, Mean ± SD | BDI, Mean ± SD | ZSRDS, Mean ± SD |
|---|---|---|---|---|---|---|
| Healthy controls | 6 M, 14 F | 34.2 ± 8.5 | 106.4 ± 16.1 | — | 4.8 ± 4.5 | 32.8 ± 5.9 |
| Depression patients | 10 M, 36 F | 33.0 ± 9.5 | 103.2 ± 14.6 | 27.1 ± 4.4 | 20.9 ± 10.0 | 46.6 ± 7.0 |

*2.2. Positive/Negative Emotion Atlas Construction*

The method for establishing the PEA and NEA is shown in Figure 2 and involves four distinct steps. Initially, data splitting and recombination are conducted, which involves timepoint segmentation based on the design of stimulus task blocks, resulting in 63 task blocks for both positive and negative emotional stimuli. Each task block comprises 5 time points of stimulus and resting-state image data. Subsequently, data undergo preprocessing and registration with the Brainnetome Atlas [29] to obtain complete time series corresponding to signals from each brain region. Following this, feature collection ensues, involves normalization of the time series using Min-Max Normalization and mean calculation, yielding matrices of positive/negative emotion features for subsequent feature selection. Feature selection is then executed utilizing the Support Vector Machine recursive feature elimination (SVM-RFE) algorithm [30] to select brain regions significantly activated by positive and negative emotional stimuli, resulting in characteristic ROIs. Subsequent data preprocessing and feature collection are then performed on the voxels within characteristic ROIs, followed by another round of SVM-RFE algorithm to filter out characteristic sub-ROIs. Given potential limitations in the extraction process, this study opts to identify external characteristic voxels exhibiting strong functional connectivity with the characteristic sub-ROIs. These voxels undergo calculation of their Pearson Correlation Coefficient, with those exceeding a correlation threshold of 0.95 being retained and integrated to finalize the PEA and NEA.

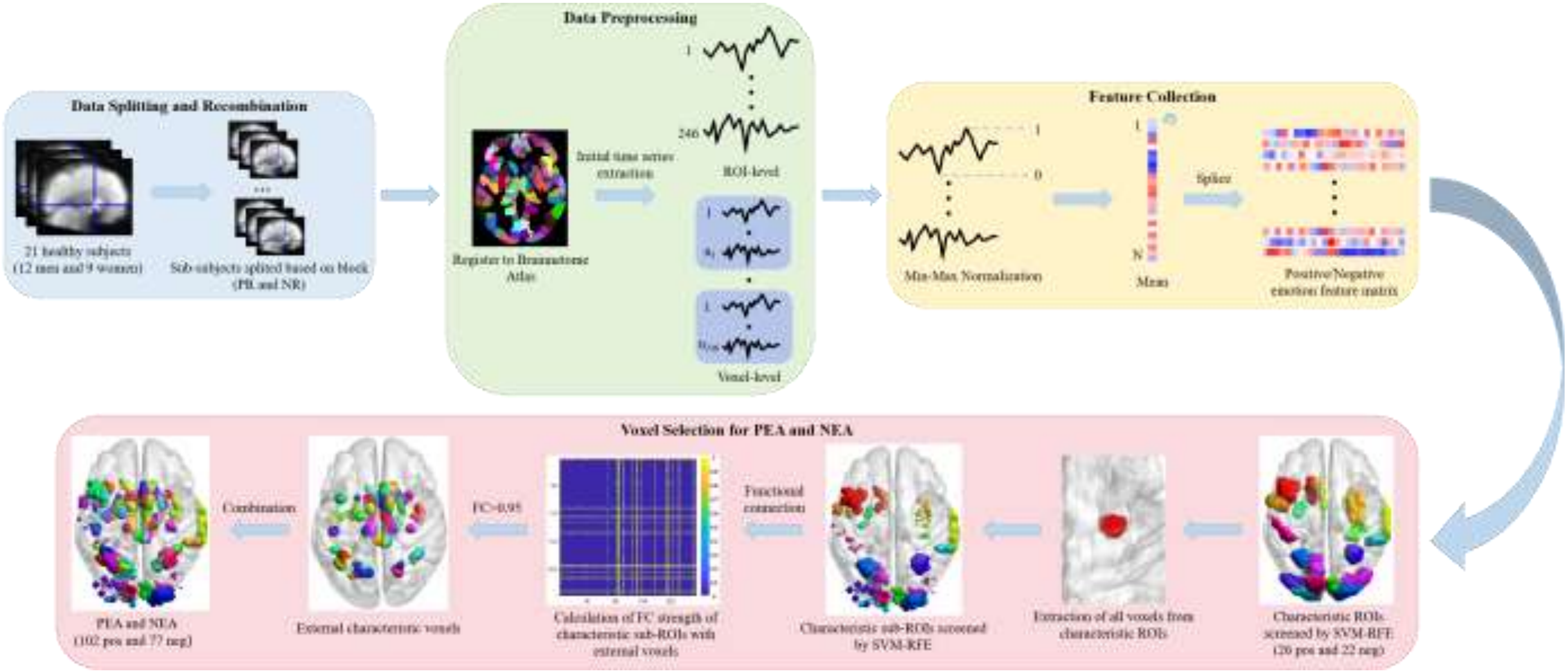

**Figure 2.** Flow chart of construction of PEA and NEA.

2.2.1. Data Splitting and Recombinations

The data sets collected for this study were divided into two categories. One type of data showed 20 s of positive emotional stimulus pictures followed by 20 s of white crosses on a black background. While the other type showed 20 s of negative emotional stimulus pictures followed by 20 s of white crosses on a black background. Considering that the BOLD signal of fMRI typically extends beyond the neural activity for 8–12 s, and recognizing the limited research on the transition from stimulus to baseline levels, post-experiment data splitting and reconstruction were performed. Specifically, in the time series of 40 s, data from 11–20 s were extracted as the data of the positive/negative emotional stimulation state, and the data of 31–40 s were extracted as the data of the recovery to the baseline state, aiming to minimize the impact of intrinsic emotional fluctuations on the study. Since it takes 2 s to scan the whole brain, each 10 s task block yielded five brain images. Following data extraction, the time series characteristics of the Positive-Rest (PR) and Negative-Rest (NR) control groups were obtained.

Given the emphasis of this paper on identifying brain regions associated with emotional function, minimizing the influence of individual differences is crucial. Therefore, six task blocks per participant were selected from the formal experiment, resulting in a total of 63 task blocks for positive and negative categories across 21 subjects. From each task block, a set of 10 brain images representing the stimulus and resting states was acquired.

2.2.2. Data Preprocessing

DPABI V8.1_240101 software [31] was used for preprocessing 20 brain images from each of the 63 task blocks. The main steps included slice timing, motion correction, spatial normalization, spatial smoothing and filtering, etc. To ensure image stability, the initial five time points were excluded from each subject's data, and corrections for septal scan discrepancies were applied using the 33rd layer as the reference. To rectify image misalignment resulting from head movements, the Friston 24 method [32] was utilized to adjust head movement parameters, with shifts exceeding 3 mm or rotations exceeding 3 degrees deemed as excessive head movements. The images were normalized using the EPI template, the voxel size was set to 3 × 3 × 3 mm$^3$, FWHM = [4 mm, 4 mm, 4 mm], and the data at the frequency of 0.01 Hz–0.1 Hz were extracted [33]. Finally, the pre-processed data were registered with Brainnetome Atlas to obtain the complete time series corresponding to the signals of each brain region for subsequent research.

### 2.2.3. Feature Collection

Following data extraction and preprocessing, each task block exhibited distinct stimulus and resting states, resulting in a data matrix of 5×246, representing 246 brain regions with data across 5 time points. Min-Max Normalization was performed on the data of each brain region in each task block, followed by the computation of the average value over the five time points. The 5×246 positive/negative emotional stimulus and resting-state matrices were converted into 1×246 matrices. Subsequently, in the comparative analyses, the datasets segregated into positive/negative emotional picture stimulation and resting-state data, yielding two sets of data matrix 126×246, o comprising 63 states of emotional stimulation and 63 corresponding resting states. For all the voxels in each brain region, the average activation value of each voxel in the positive/negative emotional stimulus state and the resting state was calculated after the Min-Max Normalization of the five time points of each voxel in the brain region. The overall process was similar to that of brain regions, and all features based on brain regions and voxels were finally obtained.

### 2.2.4. Voxel Selection for PEA and NEA

To identify the characteristic ROIs and characteristic sub-ROIs significantly activated by positive and negative emotional stimuli, a comprehensive feature selection process is essential to eliminate redundant information. In the analysis of brain region level, 246 brain regions were selected as the ROI. The average activation value of these brain regions under positive/negative emotional picture stimulation or resting state in a time series were utilized for feature selection. In the voxel-level analysis, all the voxels in a characteristic ROI were selected as the ROI, and the average activation value of the voxel in a time series was used as the feature for selecting. Positive emotional stimuli and resting state and negative emotional stimuli and resting state were utilized as two control groups in this study. The SVM-RFE algorithm was employed for feature selection of brain regions and voxels.

SVM is a generalized linear classifier, which is often used to solve binary classification problems [34]. For the input independent variable $x$ and label variable $y$, the objective function of SVM is as follows:

$$J = 1/2 \cdot \sum_{h=1}^{N} \sum_{k=1}^{N} y_h y_k \alpha_h \alpha_k (x_h \cdot x_k) - \sum_{k=1}^{N} \alpha_k \tag{1}$$

where, $0 \leq \alpha_k \leq C$ and $\sum_{k=1}^{N} \alpha_k y_k = 0$, there exists an optimal solution $\alpha_k$ for this objective function, then the decision function for input variable $x$ is as follows:

$$D(x) = w \cdot x + b \tag{2}$$

$$W = \sum_{k=1}^{N} \alpha_k y_k x_k \text{ and } b = y_k - w \cdot x_k \tag{3}$$

SVM-RFE algorithm is a feature selection method combined with SVM, which uses the weight as the ranking criterion to backward delete features, and finally obtains the optimal feature subset [35]. SVM-RFE feature selection algorithm has strong generalization ability and stable performance.

In order to make the results of feature selection more stable and robust, this study uses the SVM-RFE algorithm with 10-fold cross validation. Input two groups of training samples, $X_0 = [x_1, x_2, \ldots, x_i, \ldots, x_m]^{\mathrm{T}}$ and $y = [y_1, y_2, \ldots, y_i, \ldots, y_m]^{\mathrm{T}}$, each group of samples contains n features $s = [1,2, \ldots, n]$, For the set of all features, the feature with the smallest ranking coefficient is deleted in each iteration, until all the features are traversed, and the features that reach the maximum classification accuracy are retained as the selected features.

Each time the SVM model is trained, a weight vector $W$ is generated:

$$W = \{w_1, w_2, \dots, w_i, \dots, w_n\} \tag{4}$$

Here $w_i$ means the weight value of the *i*th ROI. Use *W* to compute the ranking criterion score for each round:

$$c_i = (w_i)^2, i = 1,2, \dots, n \tag{5}$$

(5)

Due to the randomness of this ranking score, 10-fold cross validation is introduced to obtain the ten ranking criterion scores of the current feature, which are averaged to obtain the average ranking score as shown in the following equation:

$$a_i = \sum_{j=1}^{10} c_i^j / 10, i = 1,2, \dots, n \tag{6}$$

where *j* is the *j*th fold in the 10-fold cross validation. The feature with the smallest average ranking coefficient in this round of 10-fold cross validation was deleted, and the final feature subset was obtained when the accuracy of the SVM trained classifier was no longer improved.

### 2.3. Identifying the Associated Regions in Depressed Patients Under PEA and NEA

The method for detecting associated brain regions in depressed patients under PEA and NEA is shown in Figure 3. In this study, classification validation was conducted separately across the Brainnetome Atlas template (246 brain regions), AAL template (90 brain regions) [36], Brodmann template (52 brain regions) [37], and the PEA and NEA constructed within this study. When employing SVM classification with Radial Basis Function (RBF) [38] based on CSL, we trained SVM using 10-fold cross validation to select appropriate parameter values from C (0.25, 0.5, 1, 2, 4) and gamma (0.5, 1, 2, 4, 8, 15) to enhance classification accuracy. Incorporating CSL prioritizes avoiding higher-cost errors over merely improving accuracy, as misdiagnosing depression as healthy carries potentially greater consequences in medical diagnosis than misdiagnosing healthy as depressed. Additionally, ALFF analysis was performed on depression patients and healthy controls under PEA and NEA, identifying clusters with significant differences under the two-sample *t*-test (FDR, $p < 0.01$) condition, followed by corresponding cognitive analysis.

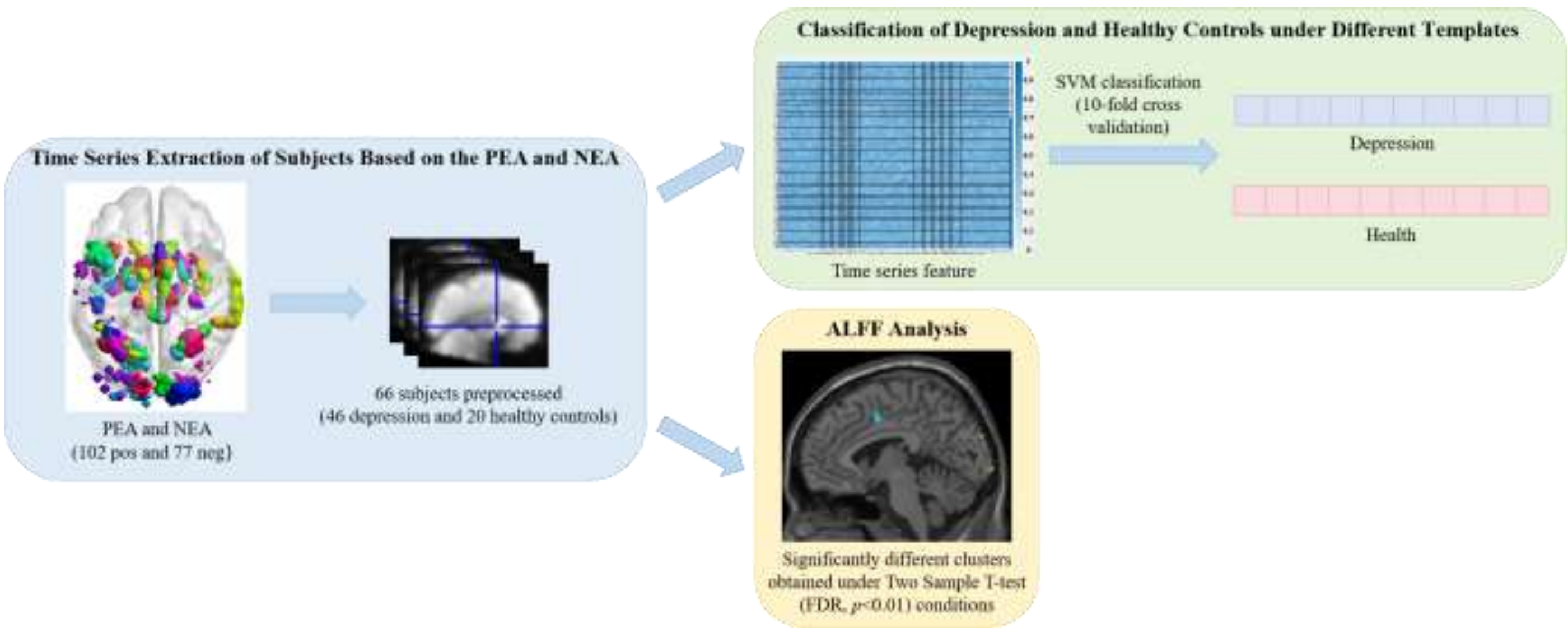

**Figure 3.** Flow chart of detection of associated brain regions in depressed patients under PEA and NEA.

## 3. Results

### *3.1. Determination of PEA and NEA in Normal Control*

During characteristic ROIs selection, a total of 246 brain regions encompassing the entire brain served as the feature set. The average activation value of brain regions under positive/negative emotional picture stimulation or resting state in a time series was utilized as the feature. The SVM-RFE algorithm was applied for feature selection, yielding emotion-related characteristic ROIs for the PR and NR groups. To identify refined emotional characteristic sub-ROIs, the SVM-RFE algorithm was iteratively employed to filter all voxels within the characteristic ROIs, thereby achieving feature dimension reduction. Subsequently, after voxel dimension extraction, refined emotional characteristic sub-ROIs for the PR and NR groups were delineated.

#### 3.1.1. Identified Positive Emotion-Associated Activation Regions

Following the implementation of the Data Splitting and Recombination procedure, 63 task blocks were generated for both positive emotional picture stimulation and resting states. Each task block corresponded to a 1×246 feature vector. The feature vectors from positive emotional stimuli and resting states were combined to form a 126×246 feature matrix for selection. Subsequently, utilizing the SVM-RFE algorithm for feature selection, 26 positive emotion-activated characteristic ROIs were identified, spanning the frontal lobe, temporal lobe, parietal lobe, insular lobe, occipital lobe, and subcortical nucleus. These brain regions are detailed in Table 2, where their importance decreases gradually from top to bottom. Notably, the activated regions are predominantly concentrated in the parietal lobe, subcortical nucleus, temporal lobe, and occipital lobe.

**Table 2.** Positive emotion-associated characteristic ROIs; Label ID is the characteristic ROI index in the Brainnetome Atlas; X, Y, Z represent the positions in MNI coordinates; L: left; R: right.

| Label ID | Gyrus | Hemisphere | MNI (X, Y, Z) |
|---|---|---|---|
| 204 | Lateral Occipital Cortex | LOcC_R_4_3 | 22, −97, 4 |
| 193 | MedioVentral Occipital Cortex | MVOcC_L_5_3 | −6, −94, 1 |
| 194 | MedioVentral Occipital Cortex | MVOcC_R_5_3 | 8, −90, 12 |
| 131 | Superior Parietal Lobule | SPL_L_5_4 | −22, −47, 65 |
| 239 | Thalamus | Tha_L_8_5 | −16, −24, 6 |
| 88 | Middle Temporal Gyrus | MTG_R_4_4 | 58, −16, −10 |
| 3 | Superior Frontal Gyrus | SFG_L_7_2 | −18, 24, 53 |
| 195 | MedioVentral Occipital Cortex | MVOcC_L_5_4 | −17, −60, −6 |
| 216 | Hippocampus | Hipp_R_2_1 | 22, −12, −20 |
| 243 | Thalamus | Tha_L_8_7 | −12, −22, 13 |
| 146 | Inferior Parietal Lobule | IPL_R_6_6 | 55, −26, 26 |
| 172 | Insular Gyrus | INS_R_6_5 | 39, −7, 8 |
| 105 | Fusiform Gyrus | FuG_L_3_2 | −31, −64, −14 |
| 108 | Fusiform Gyrus | FuG_R_3_3 | 43, −49, −19 |
| 93 | Inferior Temporal Gyrus | ITG_L_7_3 | −43, −2, −41 |
| 212 | Amygdala | Amyg_R_2_1 | 19, −2, −19 |
| 140 | Inferior Parietal Lobule | IPL_R_6_3 | 47, −35, 45 |
| 115 | Parahippocampal Gyrus | PhG_L_6_4 | −19, −12, −30 |
| 147 | Precuneus | PCun_L_4_1 | −5, −63, 51 |
| 156 | Postcentral Gyrus | PoG_R_4_1 | 50, −14, 44 |
| 138 | Inferior Parietal Lobule | IPL_R_6_2 | 39, −65, 44 |
| 21 | Middle Frontal Gyrus | MFG_L_7_4 | −41, 41, 16 |
| 214 | Amygdala | Amyg_R_2_2 | 28, −3, −20 |
| 162 | Postcentral Gyrus | PoG_R_4_4 | 20, −33, 69 |
| 57 | Precentral Gyrus | PrG_L_6_3 | −26, −25, 63 |
| 218 | Hippocampus | Hipp_R_2_2 | 29, −27, −10 |

To refine the corresponding characteristic ROIs further, this experiment conducted quadratic feature selection in the voxel dimension to delineate more detailed characteristic sub-ROIs. Figure 4 illustrates the comparison of the number of voxels in characteristic ROIs and characteristic sub-ROIs under the Brainnetome Atlas template. The results reveal a small number of retained voxels in each characteristic ROI, with some characteristic ROIs retaining only single-digit voxels exhibiting significant activation. This suggests that the fluctuations in activation value of these voxels under positive emotional stimulation significantly contribute to the activation of their respective characteristic ROIs.

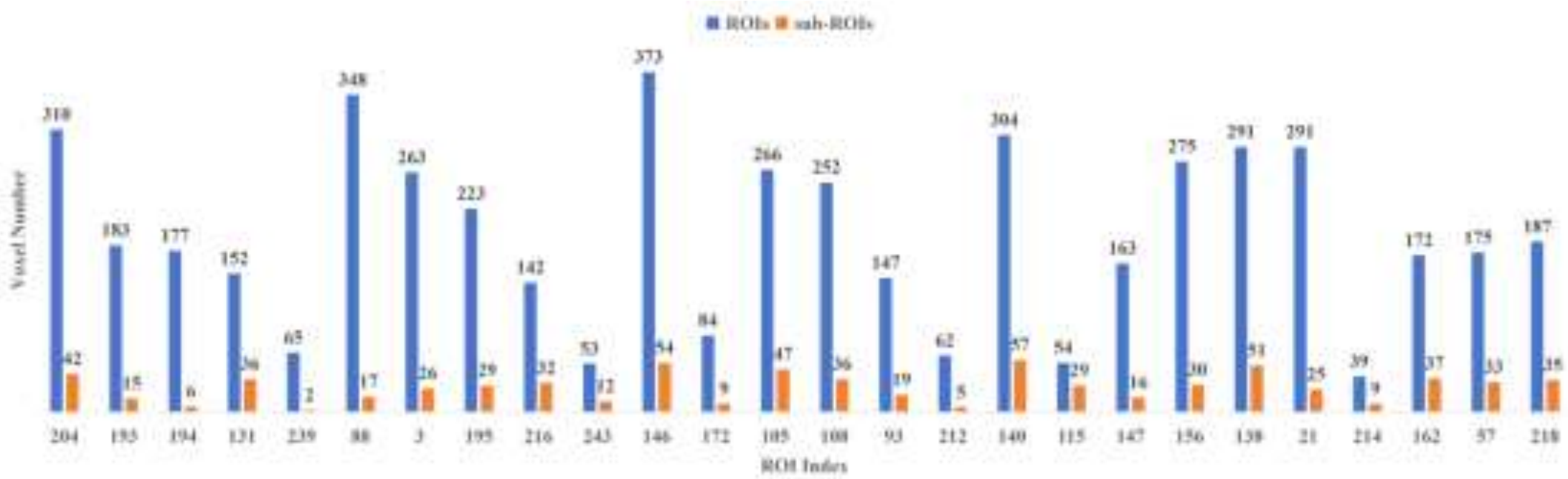

**Figure 4.** Comparison of the number of voxels in characteristic ROIs and characteristic sub-ROIs in the PR group.

The distribution of characteristic ROIs and characteristic sub-ROIs obtained by brain region and voxel feature selection in the human brain is shown in Figure 5, respectively, and it can be seen that the characteristic ROIs after the secondary feature selection are significantly reduced. The characteristic sub-ROIs serve as the basis for the subsequent construction of PEA.

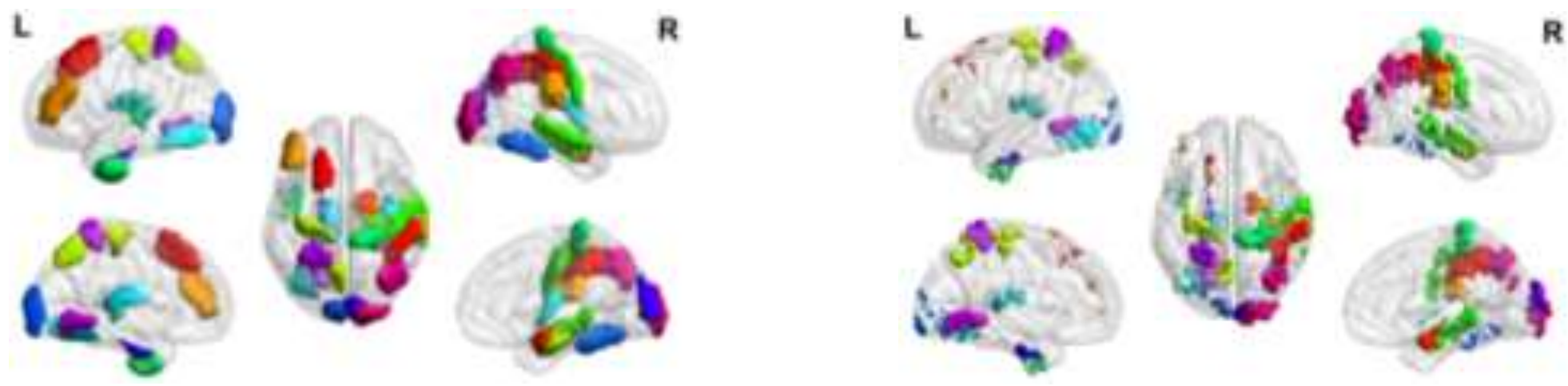

(**a**) (**b**)

**Figure 5.** (**a**) Voxel-based distribution of positive emotion-associated characteristic ROIs in the human brain; (**b**) Voxel-based distribution of positive emotion-associated characteristic sub-ROIs in the human brain.

3.1.2. Identified Negative Emotion-Associated Activation Regions

Following the implementation of the Data Splitting and Recombination procedure, 63 task blocks were generated for both negative emotional picture stimulation and resting states. Each task block corresponded to a 1×246 feature vector. The feature vectors from negative emotional stimuli and resting states were combined to form a 126×246 feature matrix for selection. Subsequently, utilizing the SVM-RFE algorithm for feature selection, 22 negative emotion-activated characteristic ROIs were identified, spanning the frontal lobe, temporal lobe, parietal lobe, insular lobe, limbic lobe, occipital lobe and subcortical nucleus. These brain regions are detailed in Table 3, where their importance decreases gradually from top to bottom. Notably, the activated regions are predominantly concentrated in occipital lobe, temporal lobe and parietal lobe.

**Table 3.** Negative emotion-associated characteristic ROIs; Label ID is the characteristic ROI index in the Brainnetome Atlas; X, Y, Z represent the positions in MNI coordinates; L: left; R: right.

| Label ID | Gyrus | Hemisphere | MNI (X, Y, Z) |
|---|---|---|---|
| 203 | Lateral Occipital Cortex | LOcC_L_4_3 | −18, −99, 2 |
| 204 | Lateral Occipital Cortex | LOcC_R_4_3 | 22, −97, 4 |
| 167 | Insular Gyrus | INS_L_6_3 | −34, 18, 1 |
| 205 | Lateral Occipital Cortex | LOcC_L_4_4 | −30, −88, −12 |
| 194 | MedioVentral Occipital Cortex | MVOcC_R_5_3 | 8, −90, 12 |
| 134 | Superior Parietal Lobule | SPL_R_5_5 | 31, −54, 53 |
| 197 | MedioVentral Occipital Cortex | MVOcC_L_5_5 | −13, −68, 12 |
| 219 | Basal Ganglia | BG_L_6_1 | −12, 14, 0 |
| 230 | Basal Ganglia | BG_R_6_6 | 29, −3, 1 |
| 104 | Fusiform Gyrus | FuG_R_3_1 | 33, −15, −34 |
| 133 | Superior Parietal Lobule | SPL_L_5_5 | −27, −59, 54 |
| 151 | Precuneus | PCun_L_4_3 | −12, −67, 25 |
| 170 | Insular Gyrus | INS_R_6_4 | 39, −2, −9 |
| 206 | Lateral Occipital Cortex | LOcC_R_4_4 | 32, −85, −12 |
| 124 | posterior Superior Temporal Sulcus | pSTS_R_2_2 | 57, −40, 12 |
| 76 | Superior Temporal Gyrus | STG_R_6_4 | 66, −20, 6 |
| 207 | Lateral Occipital Cortex | LOcC_L_2_1 | −11, −88, 31 |
| 159 | Postcentral Gyrus | PoG_L_4_3 | −46, −30, 50 |
| 17 | Middle Frontal Gyrus | MFG_L_7_2 | −42, 13, 36 |
| 109 | Parahippocampal Gyrus | PhG_L_6_1 | −27, −7, −34 |
| 61 | Precentral Gyrus | PrG_L_6_5 | −52, 0, 8 |
| 177 | Cingulate Gyrus | CG_L_7_2 | −3, 8, 25 |

To refine the corresponding characteristic ROIs further, this experiment conducted quadratic feature selection in the voxel dimension to delineate more detailed characteristic sub-ROIs. Figure 6 illustrates the comparison of the number of voxels in characteristic ROIs and characteristic sub-ROIs under the Brainnetome Atlas template. The results reveal a small number of retained voxels in each characteristic ROI, with some characteristic ROIs retaining only single-digit voxels exhibiting significant activation. This suggests that the fluctuations in activation value of these voxels under negative emotional stimulation significantly contribute to the activation of their respective characteristic ROIs.

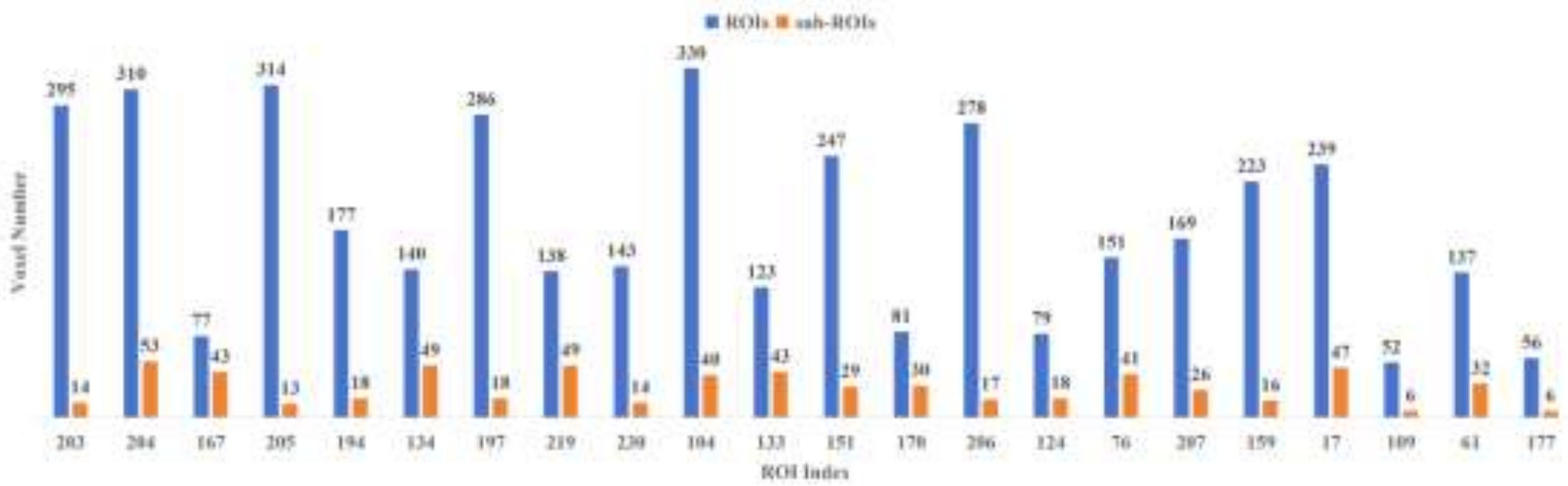

**Figure 6.** Comparison of the number of voxels in characteristic ROIs and characteristic sub-ROIs in the NR group.

The distribution of characteristic ROIs and characteristic sub-ROIs obtained by brain region and voxel feature selection in the human brain is shown in Figure 7, respectively, and it can be seen that the characteristic ROIs after the secondary feature selection are significantly reduced. The characteristic sub-ROIs serve as the basis for the subsequent construction of NEA.

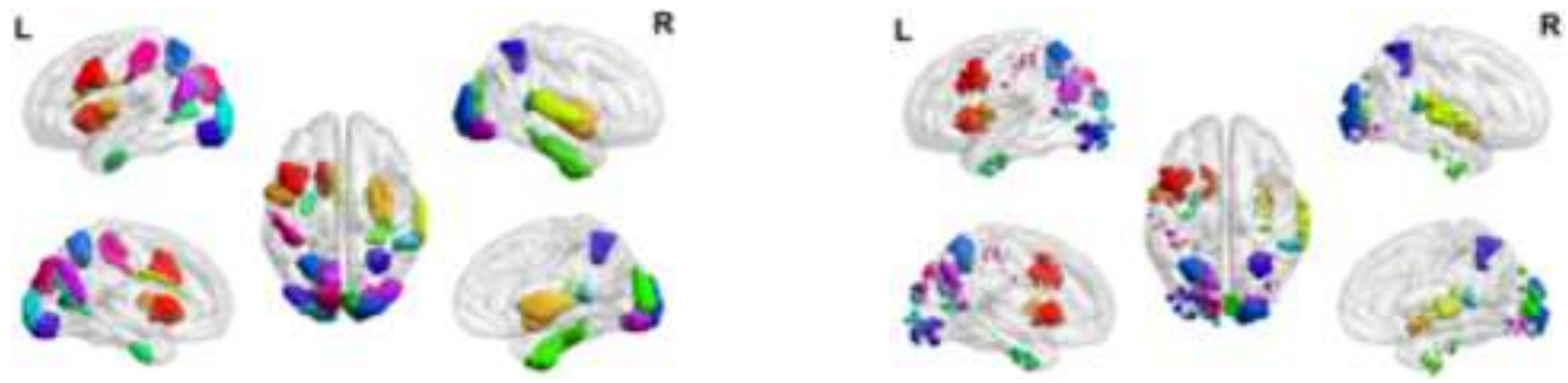

(**a**) (**b**)

**Figure 7.** (**a**) Voxel-based distribution of negative emotion-associated characteristic ROIs in the human brain; (**b**) Voxel-based distribution of negative emotion-associated characteristic sub-ROIs in the human brain.

*3.2. Results of Construction of PEA and NEA*

The number of characteristic ROIs in PR group was 26, while the number of non-characteristic ROIs was 220, and these non-characteristic ROIs contained a total of 36,398 voxels. The correlation coefficient between each voxel and the time series vector of 26 characteristic sub-ROIs was calculated. If the FC strength between each voxel and a characteristic sub-ROI was greater than 0.95, the voxel was retained. After FC analysis, a total of 776 voxels were retained and these voxels were distributed in 76 brain regions. The external characteristic voxels distribution extracted using FC strength is shown in Figure 8a. Combined with the above characteristic sub-ROIs, the final PEA of this paper was obtained, involving 102 brain regions, as shown in Figure 8b.

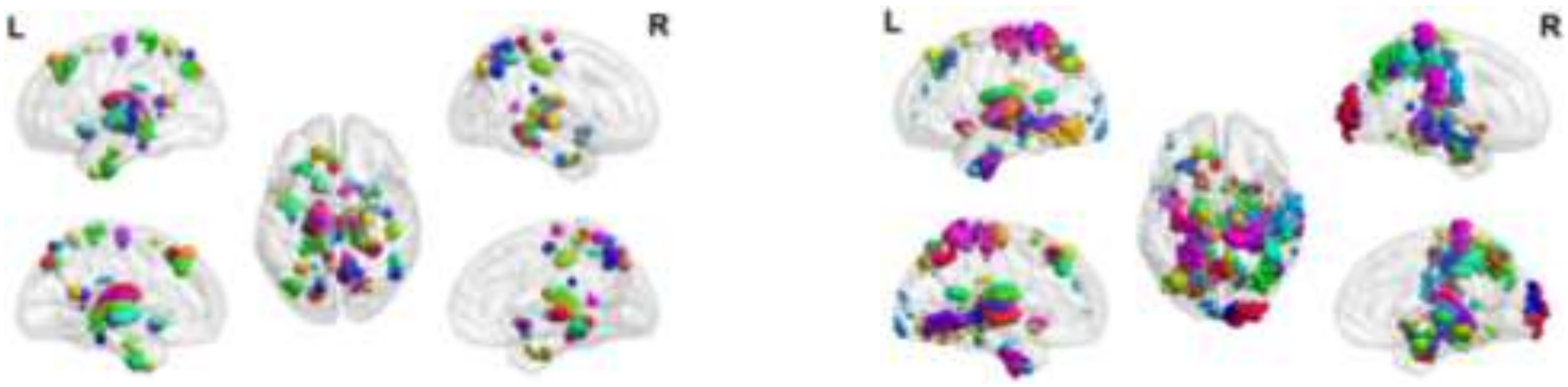

(**a**) (**b**)

**Figure 8.** (**a**) External characteristic voxels obtained by FC in the PR group; (**b**) Demonstration of PEA.

The number of characteristic ROIs in NR group was 22, while the number of non-characteristic ROIs was 224, and these non-characteristic ROIs contained a total of 37,400 voxels. The correlation coefficient between each voxel and the time series vector of 22 characteristic sub-ROIs was calculated. If the FC strength between each voxel and a characteristic sub-ROI was greater than 0.95, the voxel was retained. After FC analysis, a total of 715 voxels were retained and these voxels were distributed in 55 brain regions. The external characteristic voxels distribution extracted using FC strength is shown in Figure 9a. Combined with the above characteristic sub-ROIs, the final NEA of this paper was obtained, involving 77 brain regions, as shown in Figure 9b.

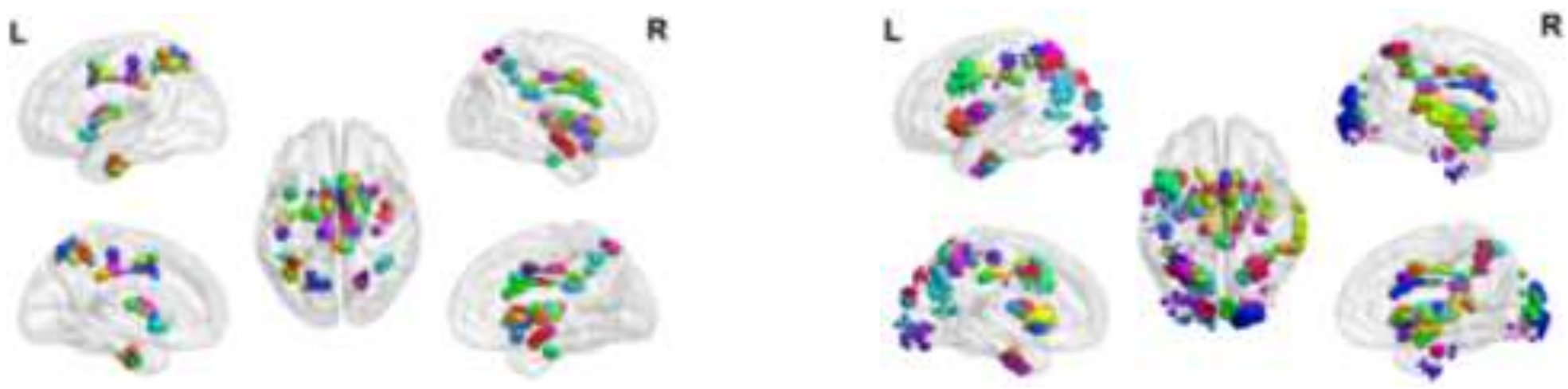

(**a**) (**b**)

**Figure 9.** (**a**) External characteristic voxels obtained by FC in the NR group; (**b**) Demonstration of NEA.

To verify the PEA and NEA constructed in this paper, we used SVM classifiers based on RBF and CSL for classification verification, and used four quality measurement indicators of Accuracy, Precision, Recall and F-score to evaluate the effectiveness of templates.

In this experiment, we evaluated the PEA and NEA constructed in this paper, and extracted the time series of PR and NR using PEA, NEA, and Brainnetome Atlas templates. As shown in Table 4, the classification accuracy of PEA and NEA both exceeded 0.80. Especially when PEA and NEA were combined to extract features for classification, all performance indicators were better than those when using PEA or NEA alone.

**Table 4.** Classification Performance Comparison of various indicators of PR and NR under different templates on positive/negative stimulus task fMRI dataset.

| Template | Accuracy | Precision | Recall | F-Score |
|---|---|---|---|---|
| Brainnetome Atlas | 64.29% | 65.21% | 64.36% | 64.31% |
| PEA | 84.65% | 84.82% | 87.69% | 85.12% |
| NEA | 81.31% | 82.88% | 80.36% | 80.86% |
| PEA + NEA | 87.27% | 87.27% | 89.26% | 87.14% |

In this experiment, we evaluated the application effects of PEA and NEA in patients with depression. We extracted time series from patients with depression and healthy controls using PEA, NEA, Brainnetome Atlas template, AAL template, and Brodmann template. As shown in Table 5, the classification accuracy of both PEA and NEA exceeds 0.70.

**Table 5.** Classification Performance Comparison of various indicators between depression and healthy control groups under different templates on resting-state depression fMRI dataset.

| Template | Accuracy | Precision | Recall | F-Score |
| --- | --- | --- | --- | --- |
| Brodmann Area | 61.79% | 64.82% | 89.98% | 75.29% |
| AAL | 62.74% | 65.64% | 90.07% | 75.86% |
| Brainnetome Atlas | 62.89% | 66.44% | 88.85% | 75.84% |
| ddPEA | 72.52% | 72.48% | 99.80% | 82.00% |
| NEA | 73.84% | 73.98% | 99.60% | 82.86% |
| PEA + NEA | 74.38% | 73.57% | 99.80% | 81.90% |

### *3.3. Associated Regions in Depressed Patients Under PEA and NEA*

Two-sample *t*-test (FDR, $p < 0.01$) was used to evaluate the difference of brain signal activation between patients with depression and healthy controls under PEA. Age and gender were included as covariates and removed to ensure the accuracy of the results. Significant differences under the PEA revealed 15 clusters, as detailed in Table 6. Brain regions implicated included the right fusiform gyrus, parahippocampal gyrus, lingual gyrus, sub-lobar, extra-nuclear, inferior parietal lobule, left parahippocampal gyrus, posterior cingulate, precuneus, precentral gyrus, thalamus, and corpus callosum. Notably, the left precentral gyrus exhibited a positive peak intensity of 2.83, while the peak intensities of other regions were negative, with the left posterior cingulate demonstrating the lowest intensity at −4.81.

**Table 6.** Clusters of brain regions with significant differences between depressed patients and healthy controls under PEA.

| Cluster | Number of Voxels | Cluster Size (mm$^3$) | Peak MNI Coordinate | Brain Regions | Peak Intensity [1] |
| --- | --- | --- | --- | --- | --- |
| 1 | 1 | 27 | 36, −48, −21 | Right Cerebrum<br>Temporal Lobe<br>Fusiform Gyrus | −2.8682 |
| 2 | 1 | 27 | 39, −42, −21 | Right Cerebrum<br>Temporal Lobe<br>Fusiform Gyrus | −2.905 |
| 3 | 2 | 54 | 15, −12, −21 | Right Cerebrum<br>Limbic Lobe | −3.4998 |
| 4 | 1 | 27 | −15, −39, −9 | Left Cerebrum<br>Limbic Lobe<br>Parahippocampa Gy-<br>rus | −2.7886 |
| 5 | 3 | 81 | 15, −33, −9 | Right Cerebrum<br>Limbic Lobe<br>Parahippocampa Gy-<br>rus<br>lingual gyrus | −3.5011 |
| 6 | 2 | 54 | −15, −24, −6 | Left Brainstem<br>Midbrain<br>Thalamus | −2.8383 |
| | | | | Medial Geniculum Body | |
| 7 | 3 | 81 | −15, −30, −3 | Left Cerebrum<br>Midbrain | −3.5216 |
| 8 | 1 | 27 | −9, −48, 6 | Left Cerebrum<br>Limbic Lobe<br>Posterior Cingulate | −4.8121 |
| 9 | 5 | 135 | 6, −42, 9 | Right Cerebrum<br>Sub-lobar<br>Extra-Nuclear<br>Corpus Callosum | −4.3618 |
| 10 | 3 | 81 | −6, −45, 15 | Left Cerebrum<br>Limbic Lobe<br>Posterior Cingulate | −3.8012 |
| 11 | 1 | 27 | 57, −33, 36 | Right Cerebrum<br>Parietal Lobe<br>Inferior Parietal Lobule | −2.7507 |
| 12 | 1 | 27 | 45, −54, 42 | Right Cerebrum<br>Parietal Lobe<br>Inferior Parietal Lobule | −2.9547 |
| 13 | 1 | 27 | −3, −57, 45 | Left Cerebrum<br>Parietal Lobe<br>Precuneus | −2.7553 |
| 14 | 1 | 27 | −9, −69, 51 | Left Cerebrum<br>Parietal Lobe<br>Precuneus | −2.9405 |
| 15 | 1 | 27 | −21, −21, 69 | Left Cerebrum<br>Frontal Lobe<br>Precentral Gyrus | 2.8264 |

[1] Peak intensity is the intensity of the maximum or minimum ALFF value in the statistically significant brain region, reflecting the extreme activity level of this brain region in the ALFF analysis, the peak intensity may be different in different brain regions and at different time points. In this study, the value of peak intensity reflects the intensity of activity in a specific brain area, with higher values indicating more intense activity in that brain area, while lower values may indicate less activity.

Two-sample *t*-test (FDR, $p < 0.01$) was used to evaluate the difference of brain signal activation between patients with depression and healthy controls under NEA. Age and gender were included as covariates and removed to ensure the accuracy of the results. Under the NEA, significant differences between the groups yielded 8 clusters, as detailed in Table 7. Implicated brain regions included the right superior temporal gyrus and middle temporal gyrus, as well as the left cuneus, middle frontal gyrus, cingulate gyrus, and superior parietal lobule. The peak intensity of the right superior temporal gyrus and middle temporal gyrus, as well as the left cuneus and middle frontal gyrus, were positive. Notably, the peak intensity of the right middle temporal gyrus was 3.11. Conversely, the peak intensity of the left cingulate gyrus, middle frontal gyrus, and superior parietal lobule was negative, with the peak intensity of the left superior parietal lobule recorded at −3.59.

**Table 7.** Clusters of brain regions with significant differences between depressed patients and healthy controls under NEA.

| Cluster | Number of Voxels | Cluster Size (mm³) | Peak MNI Coordinate | Brain Regions | Peak Intensity [1] |
|---|---|---|---|---|---|
| 1 | 1 | 27 | 69, −30, 3 | Right Cerebrum Temporal Lobe Middle Temporal Gyrus | 2.9256 |
| 2 | 3 | 81 | 66, −24, 3 | Right Cerebrum Temporal Lobe Superior Temporal Gyrus | 3.1057 |
| 3 | 1 | 27 | −12, −90, 30 | Left Cerebrum Occipital Lobe Cuneus | 2.676 |
| 4 | 1 | 27 | −42, 21, 30 | Left Cerebrum Frontal Lobe Middle Frontal Gyrus | 2.7882 |
| 5 | 1 | 27 | −3, −30, 42 | Left Cerebrum Limbic Lobe Cingulate Gyrus | −2.7923 |
| 6 | 8 | 216 | −6, −6, 48 | Left Cerebrum Limbic Lobe Cingulate Gyrus | −3.2281 |
| 7 | 3 | 81 | −51, 9, 42 | Left Cerebrum Frontal Lobe Middle Frontal Gyrus | −3.1645 |
| 8 | 5 | 135 | −9, −69, 57 | Left Cerebrum Parietal Lobe Superior Parietal Lobule | −3.5853 |

[1] Peak intensity is the intensity of the maximum or minimum ALFF value in the statistically significant brain region, reflecting the extreme activity level of this brain region in the ALFF analysis, the peak intensity may be different in different brain regions and at different time points. In this study, the value of peak intensity reflects the intensity of activity in a specific brain area, with higher values indicating more intense activity in that brain area, while lower values may indicate less activity.

## 4. Discussion

### *4.1. Positive/Negative Emotion-Associated Regions in Normal Control*

The first three activated regions of positive emotion-associated characteristic ROIs were all occipital lobe regions, including the Lateral Occipital Cortex (LOcC_R_4_3), the MedioVentral Occipital Cortex (MVOcC_L_5_3, MVOcC_R_5_3). In addition to these three, the activated regions also included the MedioVentral Occipital Cortex (MVOcC_L_5_4). Because the experimental paradigm designed in this paper requires subjects to view pictures of different emotional stimuli, the activation of emotional regions is accompanied by the activation of visual regions. The largest proportion of the 26 regions activated was in the parietal lobe, accounting for seven of them, they were the Superior Parietal Lobule (SPL_L_5_4), the Inferior Parietal Lobule (IPL_R_6_2, IPL_R_6_3, IPL_R_6_6), the Precuneus (PCun_L_4_1) and the Postcentral Gyrus (PoG_R_4_1, PoG_R_4_4). They play an important role in attention and cognitive function. By regulating the brain's attention focus and understanding and analysis ability, they make the subjects have different responses to emotional stimulus pictures [39]. Moreover, they also participate in the regulation of emotional and social behaviors, helping the brain to identify emotional expressions and participate in the understanding of social situations [40]. Meanwhile, six subcortical nuclei were activated in response to positive emotion stimulation, including the Amygdala (Amyg_R_2_1, Amyg_R_2_2), the Hippocampus (Hipp_R_2_1, Hipp_R_2_2) and the Thalamus (Tha_L_8_5, Tha_L_8_7). Among them, the amygdala and hippocampus are considered to be related to emotion, and the amygdala is mostly involved in the generation and regulation of emotion. Barrett [41] found that the amygdala is involved in predicting the threat or reward brought by emotional stimuli. While the hippocampus is associated with memory and emotional processing [42,43], the data in this paper were extracted from subjects during viewing emotional stimulus pictures, and this process may trigger short-term memory of the picture content. The thalamus is the higher center of sensation, and its activation may be related to positive emotion production [44]. Furthermore, Fusiform Gyrus (FuG_L_3_2, FuG_R_3_3) in the temporal lobe region is an important part of the ventral visual system, which processes a large number of visual and visual-related signals and is sensitive to emotional stimulus pictures [45]. The Parahippocampal Gyrus (PhG_L_6_4), as the main cortical input to the hippocampus, has an important relationship with cognition and emotion [46]. Middle Temporal Gyrus (MTG_R_4_4) and Inferior Temporal Gyrus (ITG_L_7_3), which receive information from occipital lobe input, are more advanced regions of visual processing and also function as memory regions [47]. In addition, three brain regions in the frontal lobe were significantly activated, namely the Superior Frontal Gyrus (SFG_L_7_2), Middle Frontal Gyrus (MFG_L_7_4) and Precentral Gyrus (PrG_L_6_3), which are often considered to be responsible for emotional regulation and decision-making, and can inhibit negative emotions such as anger, anxiety and fear produced by the amygdala [48]. Only one brain region in the insular lobe was significantly activated, namely the Insular Gyrus (INS_R_6_5), which is often related to the production and representation of emotions [49].

Four of the top five important activated regions for negative emotion-associated characteristic ROIs were occipital lobe regions, including Lateral Occipital Cortex (LOcC_L_4_3, LOcC_R_4_3, LOcC_L_4_4) and MedioVentral Occipital Cortex (MVOcC_R_5_3). In addition, three occipital lobe regions were included, namely Lateral Occipital Cortex (LOcC_R_4_4, LOcC_L_2_1) and MedioVentral Occipital Cortex (MVOcC_L_5_5). Similar to the positive emotion activation region, because the experimental paradigm designed in this paper requires subjects to view pictures of different emotional stimuli, the activation of emotional regions is accompanied by the activation of visual regions. Among the 22 activated regions, parietal lobe and temporal lobe regions accounted for the same proportion, and both accounted for four of them. Parietal lobe regions include the Superior Parietal Lobule (SPL_L_5_5, SPL_R_5_5), Precuneus (PCun_L_4_3), and Postcentral Gyrus (PoG_L_4_3), which are similar to the positive emotion activation region and will not be described in more detail here. Temporal lobe regions included the Superior Temporal Gyrus (STG_R_6_4), Fusiform Gyrus (FuG_R_3_1), Parahippocampal Gyrus (PhG_L_6_1) and posterior Superior Temporal Sulcus (pSTS_R_2_2). Similar to the positive emotion activation region, Watson et al. [50] found that compared with the neutral expression, emotional expression induced more significant activation in the posterior superior temporal sulcus, and the higher the expression intensity, the greater the activation. At the same time, the Basal Ganglia (BG_L_6_1 and BG_R_6_6) in the subcortical nuclei region were significantly activated, with the former often considered related to emotion, and the latter related to the production mechanism of negative emotions [51]. Furthermore, the Middle Frontal Gyrus (MFG_L_7_2) and the Precentral Gyrus (PrG_L_6_5) of the frontal lobe regions were also significantly activated, similar to the positive emotion activation region, which will not be described in more detail here. The Insular Gyrus (INS_L_6_3, INS_R_6_4) of the insular lobe region was also significantly activated, indicating that the activated regions of negative emotions were evenly and symmetrically distributed on both sides of the brain. In addition, only one brain region in the limbic lobe region was significantly activated, namely the Cingulate Gyrus (CG_L_7_2).

Papez [52] believed that emotional experience was mainly controlled by the cingulate gyrus and proposed the "Papez loop", and Kamali et al. [53] updated the original Papez loop based on this concept.

*4.2. Brain Regions in Depressed Patients Under PEA and NEA*

ALFF calculates the mean square root of the power spectrum of the signal in the low frequency range (0.01–0.08 Hz), which is used to detect the regional intensity of spontaneous fluctuations in BOLD signal [54], and the variation in the regional intensity of the BOLD signal primarily reflects changes in blood flow and blood oxygenation levels in local brain regions, which are typically associated with neural activity, especially spontaneous or task-related neural activity. Therefore, the increase of ALFF may be a sign of excessive neural activity in brain regions, while the decrease of ALFF may indicate insufficient neural activity [55]. Previous studies have reported that machine learning models trained with ALFF features have good performance in identifying patients with depression and predicting antidepressant efficacy [56,57].

In this study, we used a two-sample *t*-test (FDR, $p < 0.01$) to assess the difference in ALFF signaling activation during PEA and NEA between depressed patients and healthy controls. This method was chosen based on its widespread use and endorsement in neuroimaging studies, as demonstrated in the study by Yan et al. [58]. Among the many statistical correction methods, we preferred FDR correction because it was more stringent in controlling the false positive rate. The significance level of $p < 0.01$ was chosen to provide a conservative threshold to ensure the reliability and statistical power of the findings.

Under the PEA, depressed patients in this study had significantly lower ALFF values than healthy controls in the right fusiform gyrus, which is generally considered to be related to the storage and recognition of faces. Depressed patients are prone to misunderstand other people's facial expressions, which may be related to weaker activation. Zhang et al. [59] found that when the spontaneous brain activity of the fusiform gyrus is abnormal, patients with depression may have reduced recognition and memory of facial features, along with certain deviations in language understanding, leading them to have negative cognition in study and life. Additionally, Reynolds et al. [60] found that the functional connectivity of the right fusiform gyrus in patients with mild cognitive impairment was abnormal, which may lead to memory defects, hallucinations and emotional disorders. Therefore, the dysfunction of the fusiform gyrus may be the neurophysiological basis for patients with depression being more prone to negative emotions. Lingual gyrus is located in the visual system and plays an important role in integrating visual information, introverted sensation and stimulation [61,62]. Consistent with the research of other scholars, the ALFF value of the right lingual gyrus in patients with depression in this study was also significantly lower than that of the healthy control group. Lee et al. [63] found that the ALFF value of the bilateral lingual gyrus decreased in depressed patients with anxiety symptoms, and the gray matter connectivity of the right lingual gyrus changed abnormally in these patients. Jing et al. [64] found that compared with the healthy control group, the ALFF/fALFF value of the left lingual gyrus of patients with depression was decreased. The inferior parietal lobule is involved in many functions such as attention, sensation and spatial information integration [65]. Wang et al. [66] found that the ALFF value of the bilateral inferior parietal lobule in patients with first-episode depression decreased, with the significantly reduced region being the right inferior parietal lobule cortex. This is also consistent with the results of this study, which found that the ALFF value of the right inferior parietal lobule in patients with depression was significantly lower than that of the healthy control group. As the core brain region of the default mode network, the precuneus is related to many high-level cognitive functions and is responsible for self-related cognitive activities, such as collecting information and evaluating external emotional stimuli [67]. Sendi et al. [68] conducted a brain network study on depression and found that when there was abnormal activity in the precuneus, patients would have cognitive disorders such as inattention, active negative thoughts, and negative emotional rumination. This study found that the ALFF value of the left precuneus in patients with depression was significantly lower than that of the healthy control group. This may be related to the clinical manifestation of repeated introspection in depressed patients. Thalamus is involved in the emergence of consciousness and is a neuronal transfer station for somatosensory conduction in the human body. It interacts with and influences the prefrontal-temporal lobe, prefrontal-amygdala, and prefrontal-basal ganglia, and participates in a series of cognitive and emotional processing processes [69]. Because depression is characterized by emotional and cognitive impairments, many neuroimaging and histological studies have shown that dysfunction of the thalamus and its projection cortical targets are involved in both the pathology and physiology of depression [70]. In this study, there were significant differences in the left thalamus between patients with depression and healthy controls. We can speculate that there are functional and structural abnormalities in the thalamus in patients with depression, which cause symptoms such as loss of pleasure, attention, and memory decline. The parahippocampal gyrus can collect a variety of perceptual information, process and integrate it and then transmit it to the hippocampus, which plays a pivotal role in the cognitive processing of depression [71]. Lawrence et al. [72] found that after patients with depression and healthy controls received different degrees of facial expression image stimulation such as fear, happiness and sadness, the activation response of the right parahippocampal gyrus to positive image stimulation in patients with depression was weakened, and the activation degree of the left parahippocampal gyrus was significantly positively correlated with the severity of depressive symptoms. This is partially consistent with the results of the present study, showing that the ALFF values of the parahippocampal gyrus were significantly lower in depressed patients than in healthy controls. The posterior cingulate is mainly involved in the regulation of emotions and self-awareness, while patients with depression often show emotional instability, reduced sense of self-worth, rumination and other symptoms, which are related to abnormal activity of the posterior cingulate cortex. Caetano et al. [73] found structural changes with reduced posterior cingulate volume in patients with depression, which is in line with the results of this study that the ALFF value of the left posterior cingulate in patients with depression was significantly different from that in healthy controls. The corpus callosum is the most important nerve fiber bundle in the brain, connecting the coordination and integration of the left and right cerebral hemispheres [74]. Abnormal conduction of the corpus callosum can affect the emotional coordination, control, memory and attention of the left and right brains [75]. Li et al. [76] found that compared with the healthy control group, the fractional anisotropy (FA) values of the genu and body of the corpus callosum in patients with depression were significantly reduced, and the abnormal changes of Diffusion Tensor Imaging (DTI) only appeared in the genu and body of the corpus callosum. In this study, the ALFF value of the corpus callosum in patients with depression was significantly lower than that in healthy controls. There are few studies on the relationship between depression and the corpus callosum using fMRI technology, which can be further explored and improved in the future. The precentral gyrus is related to voluntary movement. Although this study showed that the ALFF value of the left precentral gyrus in patients with depression was significantly higher than that in the healthy control group, the patients with depression showed fewer limb movement abnormalities. Therefore, the relationship between the precentral gyrus and depression needs further research and analysis.

Under the NEA, the ALFF values in the right superior temporal gyrus and middle temporal gyrus of the depressed patients in this study were significantly higher than those of the healthy controls. The superior temporal gyrus and middle temporal gyrus are not only related to mentalization ability, but also involved in the process of explaining and predicting individual behavioral ability based on autonomous beliefs, desires and emotions, as well as in the process of semantic processing and the regulation of emotional information and cognition [77,78]. Fan et al. [79] found that ALFF activity in the right superior temporal gyrus was significantly enhanced in depressed patients with suicide attempts. Guo et al. [80] found that the ALFF values of bilateral superior temporal gyrus and middle temporal gyrus were increased in patients with depression. Therefore, it can be speculated that the abnormalities of the superior temporal gyrus and middle temporal gyrus may lead to emotional disorders in patients with depression, and subsequently leading to negative cognition, depression, anxiety and other symptoms. The occipital lobe is involved in the coding and transmission of visual information in the cortex and the perception of various facial emotions [81]. The cuneus is a part of the occipital lobe and plays a core role in the neural network related to consciousness. The function of the cuneus may be related to the process of self-reflection, and it is involved in the extraction of visuospatial imagery and episodic memory. In this study, the ALFF value of the left cuneus of patients with depression was significantly higher than that of the healthy control group, Zhou et al. [82] found a decrease in ALFF values in the left cuneus after 16 weeks of treatment with escitalopram and lithium in patients with bipolar II depression, which proved the results of this experiment in disguised form. Therefore, it can be speculated that cuneus dysfunction may lead to the preference of depressed patients for negative emotional information in episodic memory, and continuous attention to negative information may lead to the aggravation of depressive symptoms [83]. The cingulate gyrus plays a key role in cognitive and emotional information management, and it has extensive connections with brain regions that regulate emotion, the emotional valence of thoughts and autonomic nerves and visceral reflexes. Gong et al. [84] found that the ALFF value of the posterior cingulate gyrus was significantly reduced in patients with MDD and bipolar disorder. In this study, it was found that the ALFF value of the left cingulate gyrus in patients with depression was significantly lower than that in healthy controls, which was consistent with the functional disorders of emotional instability and decreased decision-making ability in patients with depression. The activation of the superior parietal lobule is not only related to response inhibition, but it is also a key brain region for response inhibition control. The damage of its structure and function can cause the impairment of inhibitory control. Response inhibition is a cognitive process that removes inappropriate behavior attempts. It is an important component of executive function and plays an important role in making correct behavioral decisions to adapt to the requirements of environmental changes. In the present study, the ALFF value of the left superior parietal lobule was significantly lower in depressed patients than in healthy controls, and it can be speculated that depressed patients have weakened response inhibition function, which leads to increased impulsivity and increases the risk of suicide in depressed patients. The middle frontal gyrus is involved in cognitive processes of executive function and working memory, as well as emotional processing. In the present study, the ALFF value of the left middle frontal gyrus was significantly higher in depressed patients than in healthy controls in one part of the voxels, while it was opposite in the other part of the voxels. Cao et al. [85] found that the ALFF value of the left middle frontal gyrus in depression patients without suicidal tendencies was significantly higher than that in healthy controls. Liu et al. [86] found that patients with bipolar disorder had lower ALFF values in the left middle frontal gyrus during depression than healthy controls. Bremner et al. [87] found that the metabolism of the middle frontal gyrus was reduced in normal people with induced depression. It has been suggested that the middle frontal gyrus may be a key site in the neuropathology of depression. These studies suggest that the dysfunction of the middle frontal gyrus plays an important role in the induction of depressive emotion and the pathogenesis of depression.

### *4.3. Neural Substrates and Manifestations of Depression Symptoms*

Depression is characterized by persistent low mood, pessimism, and a loss of interest in daily activities, often accompanied by sleep disturbances (insomnia or hypersomnia) and appetite changes (leading to weight loss or gain) [88]. Patients frequently experience fatigue, decreased self-worth, self-blame, and impaired concentration and memory, with actions becoming slow or agitated [89]. Severe symptoms include recurrent thoughts of death and suicide risk [90]. The brain regions implicated in these symptoms include the prefrontal cortex (PFC), particularly the dorsolateral prefrontal cortex (DLPFC), orbital frontal cortex, and anterior cingulate cortex (ACC), which are crucial for emotional regulation, cognitive control, problem-solving, working memory, and autonomic and neuroendocrine responses [91]. The cingulate cortex, especially the subgenual and dorsal anterior cingulate cortices, is involved in the cognitive aspects of emotion and conflict resolution of emotional stimuli. The thalamus integrates sensory stimuli, emotions, and arousal, with its volume often reduced in depression. The striatum, including the caudate nucleus and putamen, is associated with reward-oriented behavior, cognitive processes, motivation, and emotional control, and its gray matter intensity may be reduced in depression [92]. The hippocampus, related to memory and complex cognitive processes, shows volume reduction linked to emotional and cognitive symptoms of depression [93]. The amygdala, involved in emotional responses, especially negative emotions, may have increased volume in depression [94]. The insula, related to emotional experience, self-reflection, and internal visceral state assessment, shows heightened response to aversive stimuli in depression [95]. The parietal lobe, part of the default mode network (DMN), is associated with internal self-focus and rumination, with DMN overactivation linked to negative emotions in depression [96]. This study observed significant differences in multiple brain regions, closely related to depression symptoms. The fusiform gyrus, involved in facial recognition and visual processing, the parahippocampal gyrus, related to memory and emotional regulation, and the lingual gyrus and cuneus, involved in visual information processing, may all be affected [97–99]. The inferior and superior parietal lobules, related to spatial perception, and the corpus callosum, linked to cognitive and emotional symptoms, may also show abnormalities [100]. The cingulate cortex's role in self-reflection and emotional regulation, the precuneus's involvement in spatial perception, and the primary motor cortex's link to slow movement in depression are noted [101,102]. The thalamus's role in emotional regulation, the superior and middle temporal gyri's involvement in auditory processing and language comprehension, and the middle frontal gyrus's link to executive dysfunction are also highlighted [103–105].

## 5. Limitations and Future Work

This study exhibits some limitations. First, due to a relatively small sample size, especially with the emotional dataset primarily consisting of graduate students, the generalizability of the research conclusions may be somewhat limited. Secondly, since the depression dataset we used is a public dataset, it lacks detailed information about patients' comorbidities, drug therapies, and non-drug therapies (such as meditation and relaxation). This may have some impact on the research results. Finally, this study is primarily based on cross-sectional data, lacking longitudinal data support, and is thus unable to verify whether the research findings remain stable or change over time. In summary, while this study provides a new perspective on the brain mechanisms of emotional regulation in depression, further research is still needed to overcome the aforementioned limitations.

In future work, we plan to deepen our research on depression from multiple aspects. First, we will explore the classification of depression subtypes based on brain region differences and analyze the emotional expression differences among these subtypes. Second, we will conduct longitudinal studies to track changes in the differential brain regions of patients from the onset of the illness to the treatment process, assessing their correlation with changes in depression symptoms, thereby revealing the evolution of these brain regions throughout the course of depression more accurately. Additionally, we will integrate multimodal data, including brain imaging data beyond fMRI, clinical presentations, psychological assessments, and genetic factors for comprehensive analysis. Finally, we will focus on the neural plasticity of these differential brain regions, investigating how intervention treatments can promote the recovery or improvement of their functions.

## 6. Conclusions

This study utilizes fMRI technology to develop PEA and NEA based on voxel methodology, facilitating the precise identification and localization of brain regions and their associated voxels linked to positive and negative emotional responses. These atlases not only enhance the cognitive understanding of human emotional states but also, through the comparison of depression patients with healthy controls, reveals significant differences in neural mechanisms between the two groups, identifying key brain regions such as the cingulate gyrus, parahippocampal gyrus, and thalamus. With further validation and refinement, this atlas is expected to be utilized for early diagnosis of depression and the assessment of treatment efficacy.

**Author Contributions:** Conceptualization, W.Z.; resources, W.Z.; data curation, W.Z.; software, Y.F. and Y.X.; supervision, N.W.; funding acquisition, W.Z., K.Z. and N.W.; validation, Y.W. and H.Y.; investigation, Y.X.; visualization, Y.F.; methodology, Y.F.; writing—original draft, Y.F.; project administration, N.W.; writing—review & editing, W.Z., N.W., L.W., H.C., H.Y., K.Z., R.T. and W.T.S. All authors have read and agreed to the published version of the manuscript.

**Funding:** This research was funded by the National Natural Science Foundation of China [grant number 31870979, 12304526], the Hong Kong Polytechnic University Faculty Reserve Fund (Project ID: P0053738) and The Hong Kong Polytechnic University Start-up Fund (Project ID: P0053210).

**Institutional Review Board Statement:** Not applicable.

**Informed Consent Statement:** Not applicable.

**Data Availability Statement:** The dataset presented in this study can be found in online repositories. The name of the repository/repositories and accession number can be found below: https://openneuro.org/datasets/ds002748/versions/1.0.5 (accessed on 06 July 2023).

**Acknowledgments:** The editorial team and anonymous reviewers have made significant revision comments to the quality improvement of this paper. Consequently, we express our gratitude to the editorial team and anonymous reviewers.

**Conflicts of Interest:** The authors have no conflicts of interests to disclose.